\documentclass[conference]{IEEEtran}
\IEEEoverridecommandlockouts
\usepackage{cite}
\usepackage{amsmath,amssymb,amsfonts}
\usepackage{algorithmic}
\usepackage{graphicx}
\usepackage{textcomp}
\usepackage{xcolor}
\usepackage{subcaption}
\def\BibTeX{{\rm B\kern-.05em{\sc i\kern-.025em b}\kern-.08em
    T\kern-.1667em\lower.7ex\hbox{E}\kern-.125emX}}
\begin{document}

\title{Residual Speech Embeddings for Tone Classification: Removing Linguistic Content to Enhance Paralinguistic Analysis}

\author{
\IEEEauthorblockN{Hamdan Al Ahbabi}
\IEEEauthorblockA{
\textit{Khalifa University}\\
Abu Dhabi, UAE \\
hamdan.alahbabi@adia.ae}
\and
\IEEEauthorblockN{Gautier Marti}
\IEEEauthorblockA{
\textit{ADIA}\\
Abu Dhabi, UAE \\
gautier.marti@adia.ae}
\and
\IEEEauthorblockN{Saeed AlMarri}
\IEEEauthorblockA{
\textit{Khalifa University}\\
Abu Dhabi, UAE \\
saeed.almarri@adia.ae}
\and
\IEEEauthorblockN{Ibrahim Elfadel}
\IEEEauthorblockA{
\textit{Khalifa University}\\
Abu Dhabi, UAE \\
ibrahim.elfadel@ku.ac.ae}
}

\maketitle

\begin{abstract}
Self-supervised learning models for speech processing, such as wav2vec2, HuBERT, WavLM, and Whisper, generate embeddings that capture both linguistic and paralinguistic information, making it challenging to analyze tone independently of spoken content. In this work, we introduce a method for disentangling paralinguistic features from linguistic content by regressing speech embeddings onto their corresponding text embeddings and using the residuals as a representation of vocal tone. We evaluate this approach across multiple self-supervised speech embeddings, demonstrating that residual embeddings significantly improve tone classification performance compared to raw speech embeddings. Our results show that this method enhances linear separability, enabling improved classification even with simple models such as logistic regression. Visualization of the residual embeddings further confirms the successful removal of linguistic information while preserving tone-related features. These findings highlight the potential of residual embeddings for applications in sentiment analysis, speaker characterization, and paralinguistic speech processing.
\end{abstract}


\section{Introduction}

Speech is a rich medium of communication that conveys not only linguistic information (\textit{what is said}) but also paralinguistic features (\textit{how it is said}), such as tone, prosody, and speaker style. These paralinguistic cues are crucial for human understanding, influencing interpretation, sentiment, and perceived intent. However, current self-supervised speech embedding models, such as wav2vec2 \cite{baevski2020wav2vec}, encode both linguistic and paralinguistic information in an entangled manner, making it challenging to extract speaker tone independently of the spoken content. This limitation hinders applications in emotion recognition, sentiment analysis, and speaker characterization, where tone and vocal style are primary signals of interest.
Existing approaches to sentiment and emotion classification from speech embeddings often rely on raw audio representations, which capture both content and tone but lack explicit disentanglement mechanisms.
While some methods attempt to infer emotion by jointly modeling speech and text \cite{yoon2018multimodal}, they do not explicitly separate content from paralinguistic style. As a result, models may focus more on the textual meaning rather than the way a sentence is spoken, leading to suboptimal performance in tone-sensitive tasks.
To address this challenge, we introduce a simple yet effective method to isolate paralinguistic features from speech embeddings. Specifically, we propose regressing speech embeddings on their corresponding text embeddings and using the residuals as a representation of vocal tone. This residual embedding captures prosodic and speaker-specific features while removing linguistic content. Through both qualitative visualization and quantitative classification experiments, we show that these residual embeddings significantly improve tone classification performance compared to raw speech embeddings.

\subsection{Contributions} This work makes the following key contributions:

\begin{itemize}
    \item We introduce residual embeddings as a method to isolate paralinguistic features from linguistic content in self-supervised speech embeddings, improving their interpretability for tone classification.
    \item We demonstrate that residual embeddings enhance tone classification accuracy across multiple self-supervised speech embedding models.
    \item We show that residual embeddings improve linear separability, allowing logistic regression to achieve performance comparable to more complex classifiers.
    \item We provide qualitative visualizations using t-SNE, confirming that residual embeddings retain tone-related information while effectively removing linguistic content.
\end{itemize}

The remainder of the paper is organized as follows. Section II provides an overview of related work in speech embeddings, paralinguistics, and disentanglement methods. Section III describes our methodology for obtaining residual embeddings. Section IV presents qualitative and quantitative evaluations of residual embeddings in tone classification tasks. Section V discusses the limitations, potential applications, and future directions of this work.

\section{Related Work}

Recent advances in self-supervised learning (SSL) \cite{mohamed2022self} for speech processing have enabled the extraction of rich speech representations \cite{baevski2020wav2vec,hsu2021hubert,chen2022wavlm} that encode both linguistic and paralinguistic features. These embeddings have significantly improved tasks such as speaker verification, emotion recognition, and speech synthesis, demonstrating their ability to capture both content and style-related cues from speech.
However, despite their effectiveness, SSL embeddings inherently entangle linguistic content with paralinguistic tone and prosody, making them suboptimal for tasks requiring explicit tone analysis. This entanglement limits their application in sentiment analysis, speaker characterization, and vocal tone interpretation. Our work addresses this gap by explicitly disentangling tone from linguistic content, enabling a finer-grained analysis of paralinguistic features—an area that has received limited attention in prior research.

\subsection{Self-Supervised Learning for Speech Representation}

Self-supervised learning (SSL) methods such as wav2vec2 \cite{baevski2020wav2vec}, HuBERT \cite{hsu2021hubert}, and WavLM \cite{chen2022wavlm} have achieved strong performance across various speech processing tasks. A comprehensive review by Mohamed et al. (2022) \cite{mohamed2022self} categorizes SSL approaches into contrastive, predictive, and generative methods, demonstrating their effectiveness in capturing rich linguistic and paralinguistic speech features without extensive labeled data.
Our work builds upon these SSL models but introduces a novel embedding transformation: We modify existing speech embeddings to remove linguistic content and enhance tone classification. Unlike prior approaches that optimize SSL embeddings for speaker verification or speech emotion recognition, we specifically focus on how residual information within these embeddings can be leveraged for explicit tone analysis. This aligns with recent studies that investigate the unintended retention of non-targeted information in speaker embeddings, as discussed in the next section.

\subsection{Residual Information in Speech Embeddings}

Residual information in deep speaker embeddings has been investigated by Stan (2023) \cite{Stan2023}, who analyzed whether speaker representations contain unintended non-speaker-related information, such as linguistic content, recording conditions, and speaking style. Their study evaluates six state-of-the-art speaker embedding models and finds that residual information persists, indicating that speaker embeddings are not fully disentangled from linguistic content. These findings suggest that residual information within speech embeddings, rather than being mere noise, could serve as a meaningful signal for downstream tasks beyond speaker identity verification.

Building on this insight, our work introduces a regression-based residual extraction approach that explicitly removes linguistic content from speech embeddings, allowing the remaining residuals to serve as a proxy for paralinguistic tone analysis. Unlike \cite{Stan2023}, which primarily assesses the extent of identity-related leakage in speaker embeddings, we take a step further by operationalizing this residual information to enhance tone classification. While \cite{Stan2023} demonstrates that speech embeddings contain both linguistic and paralinguistic features, our method actively isolates the latter and quantitatively evaluates their effectiveness in classification tasks.

\subsection{Disentangling Linguistic and Paralinguistic Features}

Several works attempt to separate different components of speech, focusing on speaker verification and synthesis. Tu et al. \cite{tu2024contrastive} use contrastive learning with VAEs to remove linguistic content for speaker verification, while Tjandra et al. \cite{tjandra2020unsupervised} employ VQ-VAEs to model content and style separately. However, these methods do not focus on tone classification. In contrast, we directly regress out linguistic content from speech embeddings, isolating paralinguistic information for tone classification.

\subsection{Self-Supervised Speech Emotion Representation Learning}

Recent works in SSL-based emotion recognition, such as \cite{morais2022speech,ma2023emotion2vec}, optimize embeddings for emotion classification but do not explicitly remove linguistic content. Morais et al. \cite{morais2022speech} fine-tune SSL embeddings for emotion recognition, while Ma et al. \cite{ma2023emotion2vec} develop emotion2vec, a self-supervised emotion representation. Unlike these approaches, we remove linguistic content via regression, enhancing tone classification without requiring additional fine-tuning.

Our method provides a novel framework for paralinguistic analysis, improving tone classification by explicitly disentangling speech embeddings.

\section{Methodology}

\begin{figure*}[t]
    \centering
    \includegraphics[width=0.95\textwidth]{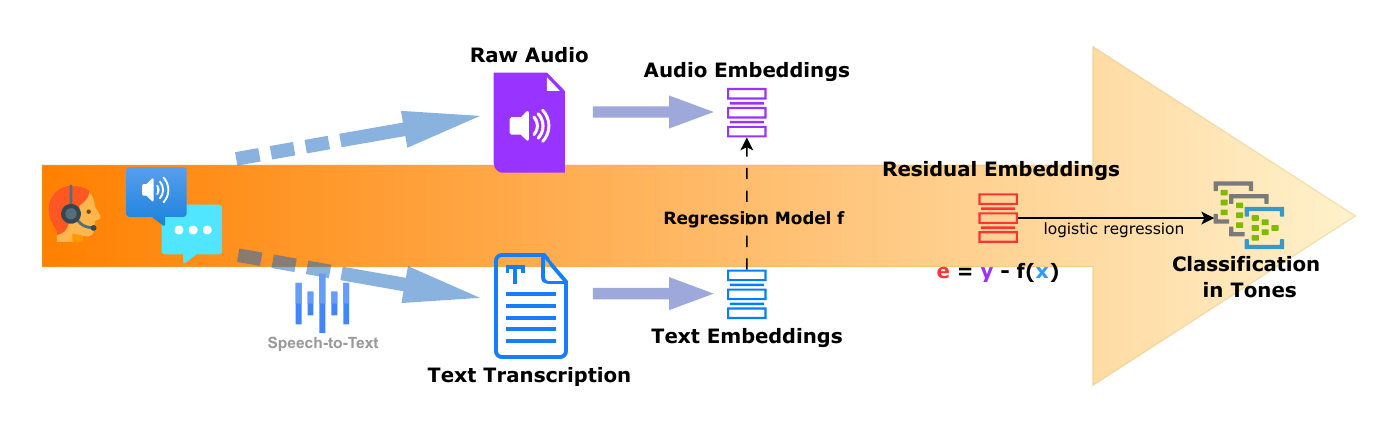}
    \caption{Overview of the regression-based residual extraction approach. Speech embeddings are regressed on text embeddings, and the residuals are used for tone classification.}
    \label{fig:methodology}
\end{figure*}

In this section, we describe our approach to disentangling linguistic content from speech embeddings and extracting residual embeddings for tone classification, illustrated in Fig.~\ref{fig:methodology}. We first introduce the dataset and feature extraction process, followed by the regression-based residual extraction method and classification framework.

\subsection{Dataset and Speech Embedding Extraction}

To systematically evaluate tone classification, we construct a synthetic dataset using \texttt{murf.ai}, a speech synthesis platform that provides precise control over vocal tone and style. The dataset consists of multiple sentences sourced from three distinct corpora (business, positive conversational, and negative conversational), each rendered in 12 tone categories, including formal, conversational, promotional, meditative, furious, and angry. All utterances are generated by the same speaker to ensure consistency in voice characteristics while isolating tonal variations.

For each utterance, we extract:
\begin{itemize}
    \item Speech embeddings \( E_s \): Extracted using wav2vec2 \cite{baevski2020wav2vec};
    \item Text embeddings \( E_t \): Extracted using OpenAI's text-embedding-ada-002 model \cite{openai2022textembed};
    \item Residual embeddings \( R \): Extracted by removing linguistic content through regression.
\end{itemize}

By leveraging synthetic speech, we ensure that tone variations remain independent of lexical differences, allowing a controlled evaluation of tone classification.

\subsection{Regression-Based Residual Extraction}

Self-supervised speech embeddings encode both linguistic and paralinguistic features, making them suboptimal for tasks that rely solely on tone, prosody, or vocal style. To explicitly remove linguistic content, we introduce a regression-based residual extraction method.

Given a speech utterance \( S \), we extract a corresponding speech embedding \( E_s \) using a self-supervised model such as wav2vec2 \cite{baevski2020wav2vec}. Simultaneously, we obtain a text-based embedding \( E_t \) of the corresponding transcription using OpenAI's text-embedding-ada-002 model \cite{openai2022textembed}. Our approach then models \( E_s \) as a function of \( E_t \) and extracts the residuals:

\begin{equation}
    \hat{E}_s = f(E_t) + R
\end{equation}

where \( f(E_t) \) is a regression function trained to predict the speech embedding from the text embedding, and \( R \) represents the residual embedding. Since \( f(E_t) \) captures the linguistic content, \( R \) retains only the paralinguistic information.

We parameterize \( f(E_t) \) as a linear model:

\begin{equation}
    f(E_t) = W E_t + b
\end{equation}

where \( W \) and \( b \) are learned regression parameters.

To estimate \( W \) and \( b \), we use ridge regression, optimized to minimize the mean squared error (MSE) between the predicted and actual speech embeddings:

\begin{equation}
    \mathcal{L} = \| E_s - f(E_t) \|_2^2 + \lambda \| W \|_2^2
\end{equation}

where \( \lambda \) is the regularization parameter that prevents overfitting. Ridge regression is chosen over standard linear regression to mitigate overfitting due to the high-dimensional nature of speech embeddings while preserving all learned coefficients.

\subsection{Tone Classification with Residual Embeddings}

The extracted residual embeddings \( R \) are then used for tone classification. We employ a logistic regression classifier as the primary method, where the goal is to assign each utterance to one of \( C \) tone categories (e.g., conversational, angry, formal, promotional).

To assess whether the information contained in the embeddings is predominantly linear, we compare the classification performance on both raw speech embeddings and residual embeddings using:

\begin{itemize}
    \item Logistic Regression (linear model) – Evaluates whether a simple linear classifier is sufficient for tone classification;
    \item Random Forest (non-linear model) – Tests whether additional non-linear interactions improve classification.
\end{itemize}

This comparison serves two purposes: (1) determining whether tone information in speech embeddings is naturally linearly separable, and (2) assessing whether the residual embeddings preserve or enhance tone classification performance under linear models.

\section{Experiments and Results}

\begin{figure*}[!t]
    \centering
    \begin{subfigure}{0.32\textwidth}
        \includegraphics[width=\linewidth]{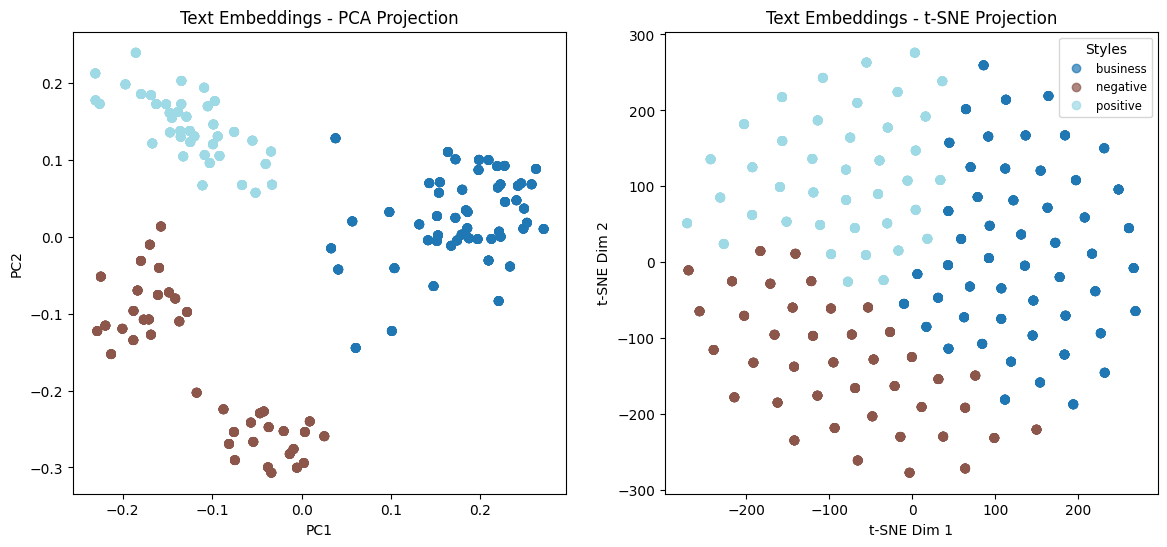}
        \caption{Text Embeddings - PCA and t-SNE}
        \label{fig:text_embeddings}
    \end{subfigure}
    \hfill
    \begin{subfigure}{0.32\textwidth}
        \includegraphics[width=\linewidth]{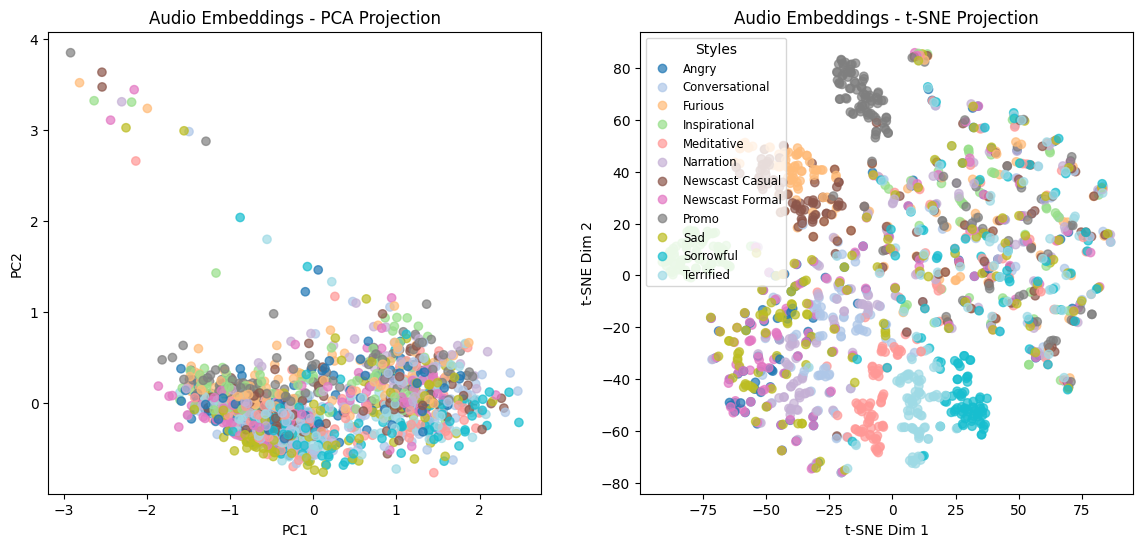}
        \caption{Audio Embeddings - PCA and t-SNE}
        \label{fig:audio_embeddings}
    \end{subfigure}
    \hfill
    \begin{subfigure}{0.32\textwidth}
        \includegraphics[width=\linewidth]{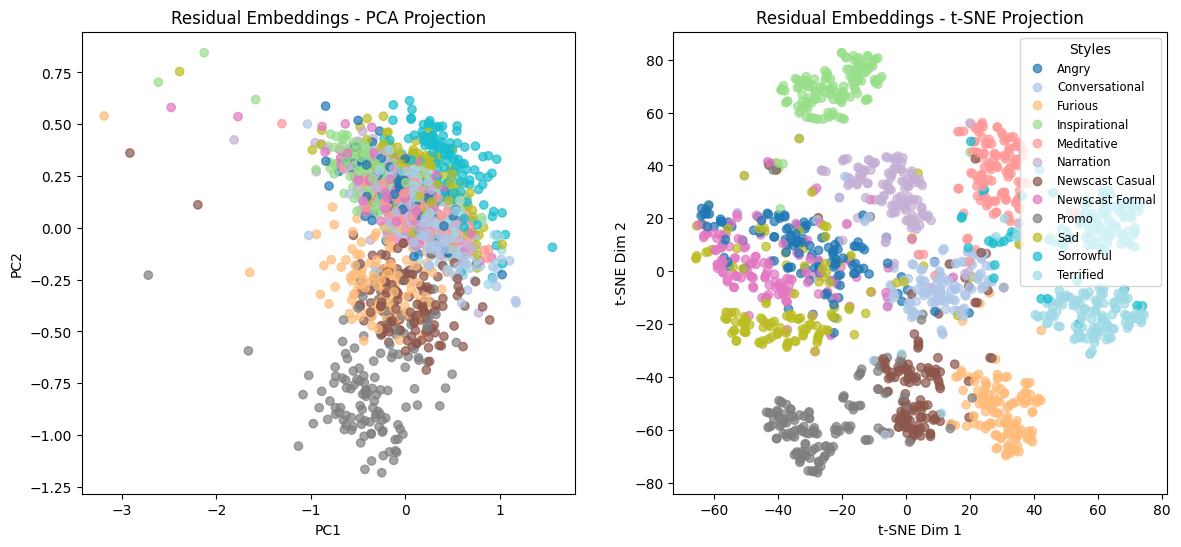}
        \caption{Residual Embeddings - PCA and t-SNE}
        \label{fig:residual_embeddings}
    \end{subfigure}
    \caption{Visualization of embeddings using PCA and t-SNE projections. (a) Text embeddings show clear separation of textual styles. (b) Audio embeddings exhibit entanglement of linguistic and paralinguistic features. (c) Residual embeddings display improved tone separability, demonstrating successful disentanglement of linguistic content.}
    \label{fig:embedding_visualization}
\end{figure*}

In this section, we evaluate the effectiveness of our regression-based residual extraction approach in improving tone classification. We conduct quantitative experiments to compare classification performance on raw speech embeddings and residual embeddings. Additionally, we visualize the extracted embeddings (Fig.~ \ref{fig:embedding_visualization}) to assess whether linguistic content has been successfully removed.

\subsection{Experimental Setup}

We evaluate our method using the synthetic dataset described in Section III, consisting of 1,584 samples derived from 52 business-related sentences, 40 positive conversational sentences, and 40 negative conversational sentences, each spoken in 12 distinct tones. We compare tone classification performance using raw speech embeddings extracted from wav2vec2 and residual embeddings obtained by regressing out linguistic content with our proposed method.

To assess whether tone information is predominantly linear, we use logistic regression as a linear classifier and random forest as a non-linear model capturing higher-order interactions. All models are trained with the same train-test split, and hyper-parameters are optimized via cross-validation to ensure robustness.

\subsection{Tone Classification Performance}

Table \ref{tab:classification_results} reports classification accuracy for raw and residual embeddings. Logistic regression achieves higher accuracy with residual embeddings, suggesting improved linear separability of tone categories. The performance gap between logistic regression and random forest also narrows when using residual embeddings further confirming that tone information is preserved in a more structured manner.

\begin{table}[htbp]
\caption{Classification Performance Comparison}
\begin{center}
\begin{tabular}{|l|c|c|c|}
\hline
\textbf{Model} & \textbf{Accuracy} & \textbf{F1 Score} & \textbf{AUC-ROC} \\
\hline
Logistic Regression (Text) & 0.00 & 0.00 & 0.12 \\
Logistic Regression (Audio) & 0.89 & 0.89 & 0.99 \\
Logistic Regression (Residual) & 0.94 & 0.94 & 1.00 \\
\hline
Random Forest (Text) & 0.00 & 0.00 & 0.10 \\
Random Forest (Audio) & 0.80 & 0.80 & 0.97 \\
Random Forest (Residual) & 0.92 & 0.92 & 1.00 \\
\hline
\end{tabular}
\label{tab:classification_results}
\end{center}
\end{table}

To ensure our findings are not specific to wav2vec2, we evaluate additional embeddings—HuBERT \cite{hsu2021hubert}, WavLM \cite{chen2022wavlm}, and Whisper \cite{radford2023robust}. As shown in Table \ref{tab:classification_summary}, the pattern remains consistent across models, reinforcing the effectiveness of residual embeddings in enhancing tone classification.

\begin{table}[htbp]
\caption{Classification Accuracy Comparison Across Models and Embeddings}
\begin{center}
\begin{tabular}{|l|c|c|c|c|}
\hline
\textbf{Model} & \multicolumn{2}{c|}{\textbf{LogReg Acc.}} & \multicolumn{2}{c|}{\textbf{RF Acc.}} \\
\cline{2-5}
 & \textbf{Audio} & \textbf{Residual} & \textbf{Audio} & \textbf{Residual} \\
\hline
HuBERT (Business)  & 0.32  & 0.87  & 0.07  & 0.86  \\
HuBERT (Sentiment) & 0.45  & 0.79  & 0.11  & 0.76  \\
\hline
WavLM (Business)   & 0.83  & 0.98  & 0.49  & 1.00  \\
WavLM (Sentiment)  & 0.74  & 0.95  & 0.42  & 0.94  \\
\hline
Whisper (Business) & 0.90  & 0.97  & 0.57  & 0.96  \\
Whisper (Sentiment)& 0.88  & 0.94  & 0.60  & 0.90  \\
\hline
\end{tabular}
\label{tab:classification_summary}
\end{center}
\end{table}

These results confirm that linguistic content in speech embeddings interferes with tone classification and that our regression-based residual extraction effectively isolates paralinguistic information across various embedding models.

\subsection{Visualizing Residual Embeddings}

To further validate our approach, we use PCA and t-SNE to visualize the distribution of embeddings before and after residual extraction (Fig.~\ref{fig:embedding_visualization}). These projections illustrate how linguistic content influences speech embeddings and how our residual extraction technique enhances tone separation.


The visualizations reveal that text embeddings (Fig.~\ref{fig:text_embeddings}) exhibit clear separation according to textual style, distinguishing business-related content from positive and negative conversational sentences. This confirms that they primarily capture linguistic structure, as expected. In contrast, audio embeddings (Fig.~\ref{fig:audio_embeddings}) display significant entanglement of linguistic and paralinguistic features, making it difficult to isolate tone information. However, residual embeddings (Fig.~\ref{fig:residual_embeddings}) show improved separability among tone categories, indicating that linguistic content has been effectively removed. These results suggest that our method successfully disentangles paralinguistic features, making them more suitable for tone classification.

\section{Conclusion and Future Directions}

Our results demonstrate that removing linguistic content from speech embeddings significantly enhances tone classification accuracy. By isolating paralinguistic features, residual embeddings consistently outperform raw speech embeddings, even with simple classifiers like logistic regression, highlighting their well-structured and separable nature. Furthermore, our experiments show that this improvement holds across multiple self-supervised speech embeddings, including wav2vec2, HuBERT, WavLM, and Whisper, indicating the robustness of our approach regardless of the underlying speech representation.

While our study is based on a controlled single-speaker synthetic dataset, real-world applications require generalization across multiple speakers with variations in pitch, intonation, and accent. Future work should extend this approach to spontaneous conversational speech and evaluate its robustness in more diverse datasets. Additionally, we aim to develop a Discrepancy Index to quantify mismatches between spoken tone and textual sentiment, which could have applications in sarcasm detection, persuasion modeling, and deception analysis. Our findings lay the foundation for improved tone analysis in AI-driven speech applications, including sentiment analysis, speaker characterization, and affective computing.

\pagebreak

\bibliographystyle{IEEEtran}  
\bibliography{ref}  

\begin{thebibliography}{10}
\providecommand{\url}[1]{#1}
\csname url@samestyle\endcsname
\providecommand{\newblock}{\relax}
\providecommand{\bibinfo}[2]{#2}
\providecommand{\BIBentrySTDinterwordspacing}{\spaceskip=0pt\relax}
\providecommand{\BIBentryALTinterwordstretchfactor}{4}
\providecommand{\BIBentryALTinterwordspacing}{\spaceskip=\fontdimen2\font plus
\BIBentryALTinterwordstretchfactor\fontdimen3\font minus \fontdimen4\font\relax}
\providecommand{\BIBforeignlanguage}[2]{{%
\expandafter\ifx\csname l@#1\endcsname\relax
\typeout{** WARNING: IEEEtran.bst: No hyphenation pattern has been}%
\typeout{** loaded for the language `#1'. Using the pattern for}%
\typeout{** the default language instead.}%
\else
\language=\csname l@#1\endcsname
\fi
#2}}
\providecommand{\BIBdecl}{\relax}
\BIBdecl

\bibitem{baevski2020wav2vec}
A.~Baevski, Y.~Zhou, A.~Mohamed, and M.~Auli, ``wav2vec 2.0: A framework for self-supervised learning of speech representations,'' \emph{Advances in neural information processing systems}, vol.~33, pp. 12\,449--12\,460, 2020.

\bibitem{yoon2018multimodal}
S.~Yoon, S.~Byun, and K.~Jung, ``Multimodal speech emotion recognition using audio and text,'' in \emph{2018 IEEE spoken language technology workshop (SLT)}.\hskip 1em plus 0.5em minus 0.4em\relax IEEE, 2018, pp. 112--118.

\bibitem{mohamed2022self}
A.~Mohamed, H.-y. Lee, L.~Borgholt, J.~D. Havtorn, J.~Edin, C.~Igel, K.~Kirchhoff, S.-W. Li, K.~Livescu, L.~Maal{\o}e \emph{et~al.}, ``Self-supervised speech representation learning: A review,'' \emph{IEEE Journal of Selected Topics in Signal Processing}, vol.~16, no.~6, pp. 1179--1210, 2022.

\bibitem{hsu2021hubert}
W.-N. Hsu, B.~Bolte, Y.-H.~H. Tsai, K.~Lakhotia, R.~Salakhutdinov, and A.~Mohamed, ``Hubert: Self-supervised speech representation learning by masked prediction of hidden units,'' \emph{IEEE/ACM transactions on audio, speech, and language processing}, vol.~29, pp. 3451--3460, 2021.

\bibitem{chen2022wavlm}
S.~Chen, C.~Wang, Z.~Chen, Y.~Wu, S.~Liu, Z.~Chen, J.~Li, N.~Kanda, T.~Yoshioka, X.~Xiao \emph{et~al.}, ``{WavLM}: Large-scale self-supervised pre-training for full stack speech processing,'' \emph{IEEE Journal of Selected Topics in Signal Processing}, vol.~16, no.~6, pp. 1505--1518, 2022.

\bibitem{Stan2023}
\BIBentryALTinterwordspacing
A.~Stan, ``Residual information in deep speaker embedding architectures,'' \emph{Mathematics}, vol.~10, 2 2023. [Online]. Available: \url{http://arxiv.org/abs/2302.02742 http://dx.doi.org/10.3390/math10213927}
\BIBentrySTDinterwordspacing

\bibitem{tu2024contrastive}
Y.~Tu, M.-W. Mak, and J.-T. Chien, ``Contrastive speaker embedding with sequential disentanglement,'' in \emph{ICASSP 2024-2024 IEEE International Conference on Acoustics, Speech and Signal Processing (ICASSP)}.\hskip 1em plus 0.5em minus 0.4em\relax IEEE, 2024, pp. 10\,891--10\,895.

\bibitem{tjandra2020unsupervised}
A.~Tjandra, R.~Pang, Y.~Zhang, and S.~Karita, ``Unsupervised learning of disentangled speech content and style representation,'' \emph{arXiv preprint arXiv:2010.12973}, 2020.

\bibitem{morais2022speech}
E.~Morais, R.~Hoory, W.~Zhu, I.~Gat, M.~Damasceno, and H.~Aronowitz, ``Speech emotion recognition using self-supervised features,'' in \emph{ICASSP 2022-2022 IEEE International Conference on Acoustics, Speech and Signal Processing (ICASSP)}.\hskip 1em plus 0.5em minus 0.4em\relax IEEE, 2022, pp. 6922--6926.

\bibitem{ma2023emotion2vec}
Z.~Ma, Z.~Zheng, J.~Ye, J.~Li, Z.~Gao, S.~Zhang, and X.~Chen, ``emotion2vec: Self-supervised pre-training for speech emotion representation,'' \emph{arXiv preprint arXiv:2312.15185}, 2023.

\bibitem{openai2022textembed}
\BIBentryALTinterwordspacing
OpenAI, ``Openai api - text-embedding-ada-002,'' 2022. [Online]. Available: \url{https://platform.openai.com/docs/guides/embeddings}
\BIBentrySTDinterwordspacing

\bibitem{radford2023robust}
A.~Radford, J.~W. Kim, T.~Xu, G.~Brockman, C.~McLeavey, and I.~Sutskever, ``Robust speech recognition via large-scale weak supervision,'' in \emph{International conference on machine learning}.\hskip 1em plus 0.5em minus 0.4em\relax PMLR, 2023, pp. 28\,492--28\,518.

\end{thebibliography}

\end{document}